\documentclass{article}

\usepackage[final]{corl_2023} % Uncomment for the camera-ready ``final'' version.

\usepackage{graphicx}
\usepackage{booktabs}
\usepackage{caption}
\usepackage{subcaption}
\usepackage{multirow}
\usepackage{chngpage}
\usepackage{wrapfig}
\usepackage{adjustbox}
\usepackage{comment}
\usepackage{amsmath}

\usepackage{amssymb}
\usepackage{color}
\usepackage{hyperref}       % hyperlinks
\usepackage{url}    

\title{Coarse-to-Fine Keypose Sampling Transformer \\ for Robotic Manipulation}

\title{Recurrent coarse-to-fine   3D  Transformers for Robotic Manipulation}
\title{Active Recurrent   3D  Transformers for Robotic Manipulation}
\title{Spatially Adaptive Computation in 3D  Transformers for Multi-Task Robotic Manipulation}
\title{Spatially Adaptive Computation in 3D Keypose Transformers for Robotic Manipulation}
\title{Coarse-to-Fine Recurrent 3D  Keypose  Transformers for Robotic Manipulation}
\title{Continuous Resolution 3D  Keypose  Transformers for Robotic Manipulation}
\title{Coarse-to-Fine  Robot Pose 3D Transformers for  Manipulation}
\title{Robot keypose prediction as 3D detection: Coarse-to-Fine  Robot Pose 3D Transformers for  Manipulation}
\title{Infinite 3D Resolution in 3D Point Transformers for Robotic Manipulation}
\title{Robot Pose Prediction as Infinite Resolution 3D Detection with  Coarse-to-Fine 3D Point Transformers}

\title{Robot Pose Prediction as Infinite Resolution 3D Detection for Robot Manipulation}
\title{Robot Pose Prediction as Infinite Resolution 3D Detection-Action Transformers for Robot Manipulation}
\title{Infinite Resolution 3D Detecting Action Transformers for Robot Manipulation}
\title{Infinite Resolution 3D  Gripper Pose Detection Transformers for Robot Manipulation}
\title{Act3D: Action Prediction as Infinite Resolution 3D Detection for Robot Manipulation}
\title{Act3D: Action Prediction as 3D  Detection for Robot Manipulation}
\title{Act3D: Action Prediction with 3D Vision-language  Transformers for Robot Manipulation}
\title{Act3D: 3D Vision-language  Transformers for Robot Manipulation}
\title{Act3D: An Infinite Resolution Action Detection Transformer for Robotic Manipulation} 

\title{Act3D: Coarse-to-Fine 3D Feature Fields for Multi-Task Robotic Manipulation}

\title{Act3D: 3D Feature Field Transformers for Multi-Task Robotic Manipulation} 

\title{Act3D:  3D Feature Field Transformers for Multi-Task Robotic Manipulation} 

\newcommand{\rebuttal}[1]{{\color{black}{#1}}}

\newcommand\blfootnote[1]{%
  \begingroup
  \renewcommand\thefootnote{}\footnote{#1}%
  \addtocounter{footnote}{-1}%
  \endgroup
}

\author{Theophile Gervet$^{*, 1}$
 \hspace{8px} Zhou Xian$^{*, 2}$ \hspace{8px} Nikolaos Gkanatsios$^{2}$ \hspace{8px} Katerina Fragkiadaki$^{1}$\vspace{3pt}\\
$^1$Machine Learning Department \hspace{6px} $^2$Robotics Institute\vspace{1pt}\\
School of Computer Science\vspace{1pt}\\
Carnegie Mellon University\vspace{3pt}\\
  \texttt{\{tgervet, xianz1, ngkanats, katef\}@cs.cmu.edu}\vspace{10pt}\\
\url{act3d.github.io}\vspace{-10pt}
}

\newcommand{\model}{\text{Act3D}}

\begin{document}

\maketitle

%===============================================================================

\begin{abstract}
3D perceptual representations are well suited for robot manipulation as they easily encode occlusions and simplify spatial reasoning. 
Many manipulation tasks require high spatial precision in end-effector pose prediction, which typically demands high-resolution 3D feature grids that are computationally expensive to process.    
As a result,  most manipulation policies operate directly in 2D, foregoing 3D inductive biases. 
In this paper, we introduce \model{}, a manipulation policy transformer that represents the robot's workspace using a  3D feature field with adaptive resolutions dependent on the task at hand. The model %learns to predicts continuous resolution 3D action maps through spatially adaptive attention operations. \model{} 
 %by  % and predicts 6 DoF robot keyposes through 3D action maps. %uses spatially-adaptive computation to predict continuous resolution 3D action maps. 
%prediction using spatially-adaptive computation. 
lifts 2D pre-trained features to 3D using sensed depth, and attends to them to compute  features for sampled 3D points. %continuous resolution 3D feature and action maps  for diverse manipulation tasks.  
%It does so by iteratively
It samples 3D point grids in a coarse to fine manner, featurizes them using relative-position attention, and selects where to focus  the next round of point sampling. In this way, it efficiently computes 3D action maps of high spatial resolution.
% using pretrained 3D features from  with adaptive spatial computation.
%It lifts 2D foundational features to 3D using sensed depth 
%3D feature clouds unprojected from one or more camera views,
%and computes a 3D action map using coarse to fine 3D point sampling. 
%and featuri
%iteratively samples 3D point grids in free space in a coarse-to-fine manner, featurizes them using relative spatial attention to the physical feature cloud, and selects the best feature point for end-effector pose prediction.
\model{} sets a new state-of-the-art in RLBench, an established manipulation benchmark,  
where it achieves 10\% absolute improvement over the previous SOTA 2D multi-view policy on 74 RLBench tasks and 22\% absolute improvement with 3x less compute over the previous SOTA 3D policy. 
%We present thorough ablations which show  relative spatial attentions, warm start from large scale vision-language pre-trained 2D backbones and weight tying across coarse-to-fine attentions iterations are critical for the model's good performance. 
%In  ablations, 
We quantify the importance of relative spatial attention, large-scale vision-language pre-trained 2D backbones, and weight tying across coarse-to-fine attentions in ablative experiments.  
 %We hope Act3D can serve as a lightweight and general multi-task manipulation policy baseline for future research o, under any   demonstration budget. 
 Code and videos are available at our project website: \url{https://act3d.github.io/}. % is a versatile baseline for any manipulation task, using any number of demonstatioons.
\vspace{-15pt}
\end{abstract}
%\method~achieves 85\% average single-task success rate on 74 RLbench tasks (13\% absolute improvement over prior SOTA), 65\% average multi-task success rate on 18 RLBench tasks with 249 variations (23\% absolute improvement with 10x less compute budget), and X\% real-world multi-task success rate with just X demos for X tasks.

\blfootnote{$^{*}$ Equal contribution}

%\keywords{6 DoF Manipulation, Implicit Behavioral Cloning, Transformers} 
%, 3D Vision, Pretraining for robotics} 
%\keywords{Detecting 6-DoF Actions, Transformers, Adaptive Computation} 

%===============================================================================

\section{Introduction}

Solutions to many robot manipulation tasks can be modeled as a sequence of 6-DoF end-effector poses (3D position and orientation). Many recent methods train neural manipulation policies to predict  3D end-effector pose sequences directly from 2D images using supervision from demonstrations~\cite{shridhar2023perceiver, guhur2023instruction, liu2022instruction, james2022coarse, brohan2022rt, sieb2020graph}. 
These methods are typically sample inefficient: they often require many trajectories to handle minor scene changes at test time and cannot easily generalize across camera viewpoints and environments, as mentioned in the respective papers and shown in our experiments.  

For a robot policy to generalize under translations, rotations, or camera view changes, it needs to be spatially equivariant \cite{zhu2022sample}, that is, to map 3D translations and rotations of the input visual scene to similar 3D translations and rotations for the robot's end-effector. 
Spatial equivariance requires predicting 3D end-effector locations through 2D or 3D action maps, depending on the action space considered, instead of regressing action locations from holistic scene or image features. 
Transporter networks~\cite{zeng2021transporter}  introduced a spatial equivariant architecture  
for 4-DoF robot manipulation: 
they re-project RGB-D input images to a top-down image and predict robot end-effector 2D translations through a top-down 2D action map. 
They showed better generalization with fewer training demonstrations than prior works. 
However, they are limited to top-down 2D worlds and 4-DoF manipulation tasks. This begs the question: how can we extend spatial equivariance in action prediction to general 6-DoF manipulation?

\begin{figure}[t!]
    \centering
    \begin{adjustbox}{center}
\includegraphics[width=1.0\textwidth,keepaspectratio]{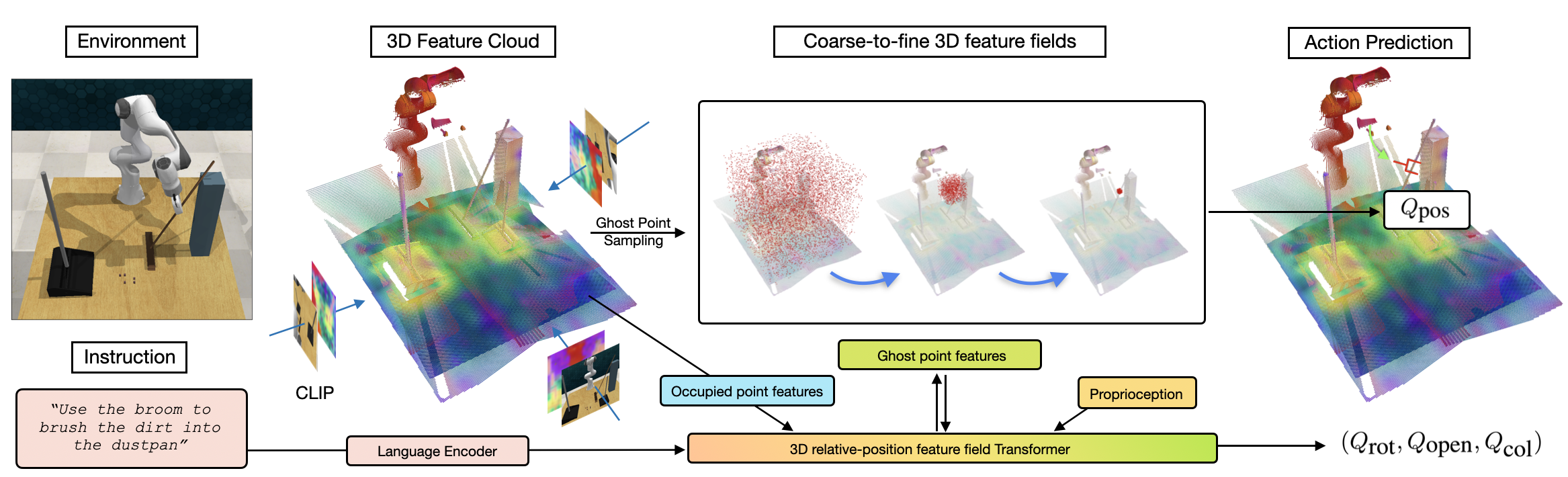}
    \end{adjustbox}    
       \caption{\textbf{\model{}}  is a language-conditioned robot action transformer that learns 3D scene feature fields of arbitrary spatial resolution via recurrent coarse-to-fine 3D point sampling and featurization using relative-position attentions.  
      \model{} featurizes multi-view RGB images with a pre-trained 2D CLIP backbone and lifts them in 3D using sensed depth. 
    It predicts 3D location of the end-effector using classification of the 3D points of the robot's workspace, which preserves spatial equivariance of the scene to action mapping. 
    % \vspace{-12pt}
    }
    \label{fig:architecture}
\end{figure}

%\rebuttal{

Developing spatially equivariant 6-DOF manipulation policies requires predicting 3D action maps by classifying 3D points in the robot's workspace as candidates for future  3D locations for the robot's end-effector. 
%Action detection  for general 6 DoF manipulation  tasks would require featurizing all 3D points of the robot's 3D workspace to classify one of them as the future end-effector 3D position. 
%Task-relevant end-effector poses  
%can be pre-grasp poses, back-off poses for articulated object interactions, or transition poses between different parts of a task and 
%can reside anywhere in the robot's 3D workspace, on points occupied by the physical scene, or in the empty space. 
Predicting high-resolution 3D action maps, necessary for  fine-grained manipulation tasks, poses a computational challenge over their 2D counterparts due to the extra spatial dimension. %, since   
%While it is straightforward to featurize 2D image pixels or occupied 3D  points, (e.g., by lifting 2D pixel features to the corresponding 3D locations using the camera extrinsics and pixel depth \cite{tung20203d},  or by using a 3D point cloud transformer~\cite{guo2021pct}) it is less clear how to  featurize points in the 3D empty space to classify one as the end-effector 3D position. 
Voxelizing the robot's 3D workspace and featurizing the 3D voxels  at high resolution is computationally demanding \cite{tung20203d}. The next end-effector pose might be anywhere in free space, which prevents the use of sparse 3D convolutions \cite{graham2015sparse,choy20194d} to selectively featurize only part of the 3D free space. 
To address this, recent work of PerAct~\cite{shridhar2023perceiver} featurizes  3D voxels using the latent set bottlenecked self-attention operation of Perceiver \cite{jaegle2021perceiver}, whose complexity is linear to the number of voxels as opposed to quadratic, as the  all-to-all self attention operations. However, it gives up on  spatial disentanglement of features due to the latent set bottleneck.  
Other methods  avoid featurizing points in 3D free space altogether and instead  regress an offset for the robot's 3D locations from a detected 2D image contact point 
~\cite{guhur2023instruction, liu2022auto, parashar2023spatial}, which again  
does not fully comply with  spatial equivariance. 
%Recent work of PerAct of bla \etal ~\cite{shridhar2023perceiver}  voxelizes a scene 3D point cloud %contributed by depth maps across one or more views 
%and featurizes \textit{all} 3D voxels (occupied or not) using the latent set bottlenecked  self-attention operation of Perceiver \cite{}. 
%PerAct requires 16 days of training on 8 V100 GPUs, more than 10 times the compute budget of our model in the same setting. 
%This inefficiency, even when projecting raw RGB, prevents them from projecting features from pre-trained 2D backbones as in CLIPort~\cite{shridhar2022cliport}.
%Many methods ~\cite{guhur2023instruction, liu2022auto, parashar2023spatial} work around this issue by avoiding featurizing points in the empty space and instead detecting a contact point, and then regressing an offset from this contact point.
%But this loses some expressiveness and doesn't fully exploit the action detection inductive bias. 

%some operate in 2D and others in 3D: 
%\model{} introduces adaptive spatial computation in a coarse-to-fine transformer policy architecture for efficient 3D keypose detection for language-conditioned multi-task robot manipulation. 
%\rebuttal{
In this paper, we introduce \model{}, a language-conditioned  transformer  for multi-task 6 DoF robot manipulation that predicts  continuous resolution 3D action maps % of infinite spatial resolution. % 
through  adaptive 3D spatial computation. \model{}  represents the scene as a continuous 3D feature field. %, that is, avoids 3D voxelization altogether and instead  
It  computes a scene-level physical 3D feature cloud by lifting features of 2D foundational models   from one or more views using sensed depth. 
%It samples 3D points in the empty workspace and featurizes them using cross-attentions to the physical 3D point features. 
It learns a 3D feature field of arbitrary spatial resolution via recurrent coarse-to-fine 3D point sampling and featurization. 
At each iteration, the model  samples 3D points in the whole workspace and featurizes them using relative spatial cross-attention~\cite{su2021roformer} to the physical 3D feature cloud. 
%In this paper, we propose \model{}, a policy transformer architecture for multi-task robot manipulation that casts end-effector keypose prediction as 3D detection. 
%Given a scene 3D feature cloud computed by ``lifting" 2D image features from one or more views using sensed depth, 
\model{} predicts 3D end-effector locations by scoring 3D point features, and then regresses the 3D orientation and  opening of the end-effector. %} % using contextualized local features around the 3D predicted location. % from the contextualized parametric query. 
At inference time, we can trade-off compute for higher spatial precision and task performance by sampling more 3D points in free space than the model ever saw at training time.

%\model uses relation position attentions, which we found critical for generalization without any need of spatial augmentations. 

%relative attentions critical  for generalization across translations 
%and handling the arbitrariness of the world coordinate system preventing overfitting 
%and also for deploying the model using more iterations at test time. 
%This scheme iteratively samples air points, first coarsely across the entire workspace then more densely near the point selected at the previous iteration, and lets them attend to the scene feature cloud at different levels of granularity.

We test \model{} in 
RLBench~\cite{james2020rlbench}, an established benchmark for learning diverse robot manipulation policies from demonstrations. We set a new state-of-the-art in the benchmark in both single-task and multi-task settings. Specifically, we achieve 
a 10\% absolute improvement over prior SOTA 
on the single-task setting introduced by HiveFormer~\cite{guhur2023instruction} with 74 tasks 
and a 22\% absolute improvement over prior SOTA 
in the multi-task setting introduced by PerAct~\cite{shridhar2023perceiver} with 18 tasks and 249 variations. 
We also validate our approach on a Franka Panda with a multi-task agent trained from scratch on 8 real-world tasks with a total of just 100 demonstrations (see Figure~\ref{fig:tasks}). 
In thorough ablations, we show the importance of the design choices of our architecture, specifically, relative spatial attention, large-scale vision-language pre-trained 2D backbones, high resolution featurization and  weight tying across coarse-to-fine attentions.
%, such as the number of coarse-to-fine stages, whether to regress a position offset, weight-tying during coarse-to-fine featurization, relative-position 3D attention, and the choice of pre-trained 2D backbone, 
%to reveal what matters for performance. 

%\rebuttal{
In summary, our contributions are:
\textbf{1.} A novel neural policy architecture for language-conditioned multi-task 6-DoF manipulation that both reasons directly in 3D and preserves locality of computation in 3D, using iterative coarse-to-fine translation-invariant attention. 
%We empirically show \model{} is more data efficient than existing 2D policy methods
%due to inductive biases of geometry, action detection, and 3D translation invariance. It is also 
%more compute-efficient at higher precision than existing 3D methods due to recurrent coarse-to-fine 3D point grid sampling and featurization instead of voxelization. 
\textbf{2.} Strong empirical results on a range of simulated and real-world tasks, outperforming the previous SOTA 2D and 3D methods on RLBench by large absolute margins, and generalizing well to novel camera placements at test time.
\textbf{3.} Thorough ablations that quantify the contribution of high-resolution features, tied attention weights, pre-trained 2D features, and relative position attention design choices. 

%===============================================================================

\section{Related Work}
% \vspace{-0.05in}
\paragraph{Learning robot manipulation from demonstrations}
Many recent work train multi-task manipulation policies that leverage Transformer architectures~\cite{shridhar2023perceiver, guhur2023instruction, liu2022instruction, brohan2022rt, shafiullah2022behavior, reed2022generalist}  to predict robot actions from video input and language instructions. 
End-to-end image-to-action policy models, such as RT-1~\cite{brohan2022rt}, GATO~\cite{reed2022generalist}, BC-Z~\cite{jang2022bc}, and  InstructRL~\cite{liu2022instruction}, directly predict 6-DoF end-effector poses from 2D video and language inputs.
They require many thousands of demonstrations to learn spatial reasoning and generalize to new scene arrangements and environments. 
Transporter networks \cite{zeng2021transporter} and their subsequent variants \cite{seita2021learning,shridhar2022cliport, gkanatsios2023energy} formulate 4-DoF end-effector pose prediction as pixel classification in 2D overhead images. 
%Their \textit{action detection} inductive bias --- parametrizing the action implicitly~\cite{florence2022implicit} by detecting end-effector poses in the scene using local features with translation and rotation equivariances~\cite{huang2022equivariant} --- 
Thanks to the spatial equivariance of their architecture, their model dramatically increased sample efficiency over previous methods that regress end-effector poses by aggregating global scene features.
However, they are limited to top-down 2D planar worlds with simple pick-and-place primitives. 
3D policy models of C2F-ARM~\cite{james2022coarse} and PerAct~\cite{shridhar2023perceiver} voxelize the robot's workspace and are trained to detect the 3D voxel that contains the next end-effector keypose. 
Spatially precise 3D pose prediction requires the 3D voxel grid to be high resolution, which comes at a high computational cost. 
C2F-ARM \cite{james2022coarse} uses a coarse-to-fine voxelization  to handle computational complexity, while PerAct~\cite{shridhar2023perceiver} uses Perceiver's latent bottleneck~\cite{jaegle2021perceiver} to avoid voxel-to-voxel self-attention operations. \model{} avoids 3D voxelization altogether and instead represents the scene as a continuous resolution 3D feature field.  
It samples 3D points in the empty workspace and featurizes them using cross-attentions to the physical 3D point features. 
\paragraph{Feature pre-training for robot manipulation}
Many 2D policy architectures bootstrap learning from demonstrations from frozen or finetuned 2D image backbones \cite{nair2022r3m,parisi2022unsurprising,jang2022bc,yenchen2021learning} to increase experience data sample efficiency.
Pretrained vision-language backbones can enable generalization to new instructions, objects, and scenes \cite{stone2023openworld,shridhar2022cliport}.  
In contrast, SOTA 3D policy models are typically trained from scratch from colored point clouds input~\cite{shridhar2023perceiver,james2022coarse, xian2021hyperdynamics}. 
\model{} uses CLIP pre-trained 2D backbones \cite{radford2021learning} to featurize 2D image views and lifts the 2D features in 3D using depth \cite{tung2019learning, harley2019learning}.
We show that 2D feature pretraining gives a considerable performance boost over training from scratch.
%\vspace{-0.05in}

\paragraph{Relative attention layers}
Relative attentions have shown improved performance in many 2D visual understanding tasks and language tasks \cite{shaw2018selfattention,liu2022swin}. 
Rotary embeddings \cite{su2022roformer} implement relative attention efficiently by casting it as an inner-product in an extended position feature space. 
In 3D, relative attention is imperative as the coordinate system is arbitrary. %; 3D feature clouds do not have the photographic biases that 2D images often have due to the regularity of image framing and 2d coordinate system placement in the center of the image. 
3D relative attentions have been used before in 3D Transformer architectures for object detection and point labelling \cite{wu2022point,yang2023swin3d}.  
We show in Section \ref{sec:experiments} that relative attentions significantly boost performance of our model. 

\section{3D Feature Field Transformers for Multi-Task Robot Manipulation}
%\vspace{-0.05in}
%\katef{ time should be either subscript or superscript, not both}
%As illustrated in Fig.~\ref{fig:architecture}, at a given timestep $t$, \model{} maps multi-view RGB-D images $O_t$, language instructions $L$, and proprioception (the current end-effector position) $Q_{\textrm{\small pos}}$ to the next best 6-DoF end-effector keypose $K_t$.
%\katef{equations for loss and for relatuve attentions we need}
%a 3D rotation and translation
The architecture of \model{} is shown in Figure~\ref{fig:architecture}. 
It is a policy transformer  that, at \rebuttal{a given} timestep $t$, predicts a  6-DoF end-effector pose from one or more RGB-D images, a language instruction, and proprioception information regarding the robot's current end-effector pose. 
%The key idea is to \emph{detect} 6 DoF end-effector poses in the robot's workspace by learning 3D perceptual representations of free space with arbitrary spatial resolution, via recurrent coarse-to-fine 3D point grid sampling and featurization. 3D point candidates (which we will call ghost points)  are sampled, featurized and scored iteratively through relative cross-attention~\cite{su2021roformer} to the physical 3D scene feature cloud, lifted from 2D feature maps of the input image views. 
Following prior work \cite{james2022q, shridhar2023perceiver, guhur2023instruction, liu2022instruction}, instead of predicting an end-effector pose at each timestep, %we instead fitting a model to predict each 6-DoF pose in the demonstration, 
we extract a set of \textit{keyposes} %$\{K_1, K_2, .., K_m\} \subset Q$ 
that capture bottleneck end-effector poses in a demonstration. % with a simple heuristic: 
A pose is a keypose if (1) the end-effector changes state (something is grasped or released) or (2) velocities approach near zero (a common occurrence when entering pre-grasp poses or entering a new phase of a task). 
The prediction problem  then boils down to  predicting the next (best) keypose action given the current observation. 
%\model{} relies purely on the current observation to predict the next action, that is, it does . 
At inference time, \model{} iteratively predicts the next best keypose and reaches it with a \rebuttal{sampling-based} motion planner, following previous works \cite{shridhar2023perceiver,guhur2023instruction}. 

%At test time, an RRT planner and inverse kinematics are used to take the robot from one keypose to the next, following previous works \cite{}.

%\xian{we need some equations to described the transformer architecture and relative 3d pos encoding}

We assume access to a dataset of $n$  demonstration trajectories.
%These demonstrations are collected by an expert with the aid of a motion-planner to reach intermediate poses. 
%\katef{let's not use Q for the action}
Each demonstration is a sequence of observations $O = \{o_1, o_2, .., o_t\}$ paired with continuous actions $A = \{a_1, a_2, .., a_t\}$ and, optionally, a language instruction $l$ that describes the task. % (in the multi-task setting). 
Each observation $o_t$ consists of  RGB-D images %($256$x$256$ resolution in our experiments) 
from one or more camera views; more details are in Appendix~\ref{sec:sim}. 
An action $a_t$ consists of the 3D position and 3D orientation (represented as a quaternion) of the robot's end-effector, its binary open or closed state, and whether the motion planner needs to avoid collisions to reach the pose:

%, and whether the motion-planner used collision avoidance to reach an intermediate pose: 
$$a = \{ a_{\mathrm{\small pos}} \in \mathbb{R}^3, a_{\mathrm{\small rot}} \in \mathbb{H}, a_{\mathrm{\small open}} \in \{0,1\}, a_{\mathrm{\small col}} \in \{0,1\} \}$$
%$$(a_{\mathrm{\small rot}}, a_{\mathrm{\small open}}, a_{\mathrm{\small col}})$$

% The set of quaternions is usually represented as H, it's a subset of R^4, open-close is binary
%\todo{are queternions $\in \mathbb{R}^4$, we need to remove that H, the open close is  not binary?}

Next, we describe the model's architecture in detail. %we go into details on the modules of \model{}. 
%$, Q_{\textrm{\small col}}\}$.

%featurization by relative cross-attention~\cite{su2021roformer} to the 3D physical scene feature cloud, followed by detection of the next best end-effector position with a detection transformer head ~\cite{carion2020end, cheng2022masked}.

%Given a 3D scene feature cloud, \model{} predicts 6 DoF end-effector keyposes through scoring 3D point features in the robot's workspace to compute 3D location, and regression to 3D rotation. 3D point candidates (which we will call ghost points)  are sampled, featurized and scored iteratively through relative attentions to the physical 3D scene feature cloud.

%Coarse-to-fine end effector pose prediction Transformers (\model{}) is a model for 6-DoF end-effector keypose prediction from one or more RGB-D images and language instructions. The key idea is to detect 6 DoF end-effector keyposes in the robot's workspace, which we'll refer to as ghost points, by learning 3D perceptual representations of empty space of arbitrary spatial resolution via recurrent coarse-to-fine 3D point grid sampling, featurization by relative cross-attention~\cite{su2021roformer} to the 3D physical scene feature cloud, followed by detection of the next best end-effector position with a detection transformer head ~\cite{carion2020end, cheng2022masked}. See Figure~\ref{fig:architecture} for an overview.

%\subsection{Demonstrations and Keyposes}

%\subsection{\model{} Architecture}

%Let us step through each component in Fig.~\ref{fig:architecture} in detail.
%\vspace{-0.05}
\paragraph{Visual and language encoder} 
Our visual encoder maps multi-view RGB-D images into a multi-scale 3D scene feature cloud. 
We use a large-scale pre-trained 2D feature extractor followed by a feature pyramid network~\cite{lin2017feature} to extract multi-scale visual tokens for each camera view. Our input is RGB-D, so each pixel is associated with a depth value. 
%After featurizing  the image, we obtain a feature map whose spatial resolution is lower than the original image. We associate every super-pixel (2D grid location) in this feature map to a depth value, by averaging the depth values of the image pixels that correspond to this super-pixel, i.e., the receptive field. Then 
We ``lift” the extracted 2D feature vectors to 3D using the pinhole camera equation and the camera intrinsics, based on their average depth.  
The language encoder featurizes instructions with a large-scale pre-trained language encoder.
We use the CLIP ResNet50~\cite{radford2021learning} visual encoder and language encoders to exploit their common vision-language feature space for interpreting instructions and referential grounding. 
Our pre-trained visual and language encoders are frozen, not finetuned, during training of Act3D.  % Please, see Appendix~\ref{sec:architecture_details} for details. 

%\todo{it is important to say that these two encoders were pretrained jointly, so one recognizes the features of the other, they were part of the same clip model}

%Note that our framework enables easily swapping the CLIP late-fusion vision and language backbones with an early fusion vision-language backbone~\cite{geng2022multimodal}.
%\vspace{-0.05in}
\paragraph{Iterative 3D point sampling and featurization} 
Our key idea is to estimate high resolution 3D action maps %\emph{detect} 6 DoF end-effector poses in the robot's workspace
by learning 3D perceptual representations of free space with arbitrary spatial resolution, via recurrent coarse-to-fine 3D point sampling and featurization. 3D point candidates (which we will call ghost points)  are sampled, featurized and scored iteratively through relative cross-attention~\cite{su2021roformer} to the physical 3D scene feature cloud, lifted from 2D feature maps of the input image views. 
%For precise and computationally tractable keypose detection, we sample, featurize and select ghost points iteratively, 
We first sample coarsely across the entire workspace, then finely in the vicinity of the ghost point selected as the focus of attention in the previous iteration, as shown in Figure \ref{fig:architecture}. The coarsest ghost points attend to a global coarse scene feature cloud, whereas finer ghost points attend to a local fine scene feature cloud. %\rebuttal{See Appendix~\ref{sec:architecture_details} for details.}

%\vspace{-0.05in}
\paragraph{Relative 3D cross-attentions} 
We featurize each of the 3D ghost points and a parametric query (used to select via inner-product one of the ghost points as the next best end-effector position in the decoder) independently through cross-attentions to the multi-scale 3D scene feature cloud, language tokens, and proprioception.
Featurizing ghost points independently, without self-attentions to one another, enables sampling more ghost points at inference time to improve performance, as we show in Section~\ref{sec:experiments}. 
Our cross-attentions use relative 3D position information and are implemented efficiently with rotary positional embeddings~\cite{su2021roformer}. The absolute locations of our 3D points are never used in our featurization, and attentions only depend on the relative locations of two features. %as a result the model is translation invariat

\paragraph{Decoding actions}
%Once ghost points and the parametric query are featurized, the detection transformer head 
We score ghost point tokens via inner product with the parametric query to select one as the next best end-effector position $a_{\mathrm{\small pos}}$. 
We then regress the end-effector orientation $a_{\mathrm{\small rot}}$ and opening $a_{\mathrm{\small open}}$, as well as whether the motion planner needs to avoid collisions to reach the pose $a_{\mathrm{\small col}}$, from the \rebuttal{last iteration} parametric query with a \rebuttal{2-layer} multi-layer perceptron (MLP).

%\vspace{-0.05in}
\paragraph{Training}
\model{} is trained supervised from input-action tuples from a dataset of manipulation demonstrations. These tuples are composed of RGB-D observations, language goals, and keypose actions $\{(o_1, l_1, k_1),(o_2, l_2, k_2), ...\}$. During training, we randomly sample a tuple and supervise \model{} to predict the keypose action $k$ given the observation and goal $(o, l)$. We supervise position prediction $a_{\mathrm{\small pos}}$ at every round of coarse-to-fine with a softmax cross-entropy loss over ghost points, rotation prediction $a_{\mathrm{\small rot}}$ with a MSE loss on the quaternion prediction, and binary end-effector opening $a_{\mathrm{\small open}}$ and whether the planner needs to avoid collisions $a_{\mathrm{\small col}}$ with binary cross-entropy losses. 
%Our loss reads:

%\begin{equation}
%\mathval{L}=\|\hat{a}_{\mathrm{\small rot}} - a_{\mathrm{\small rot}}\|_2 + \sum_ \mathbf{1}_{a_open==1} p(
%\end{equation}
%\vspace{-0.05in}
\textbf{Implementation details}
%We extract two feature maps per $256$x$256$ input image view: $32$x$32$ coarse visual tokens and $128$x$128$ fine visual tokens.
We use three ghost point sampling stages: first \rebuttal{uniformly} across the entire workspace (roughly $1$ meter cube), then \rebuttal{uniformly} in a $16$ centimeter diameter ball, and finally in a $4$ centimeter diameter ball. 
The coarsest ghost points attend to a global coarse scene feature cloud ($32$x$32$x$n_{\textrm{\small cam}}$ coarse visual tokens) whereas finer ghost points attend to a local fine scene feature cloud (the closest $32$x$32$x$n_{\textrm{\small cam}}$ out of the total $128$x$128$x$n_{\textrm{\small cam}}$ fine visual tokens). 
During training, we sample $1000$ ghost points in total split equally across the three stages. 
At inference time, we can trade-off extra prediction precision and task performance for additional compute by sampling more ghost points than the model ever saw at training time ($10,000$ in our experiments). 
We'll show in ablations in Section~\ref{sec:experiments} that our framework is robust to these hyper-parameters but tying weights across sampling stages and relative 3D cross-attention are both crucial for generalization.
%We use $2$ layers of cross-attention and an embedding size $60$ for single-task experiments and $120$ for multi-task experiments.
%Training samples are augmented with random crops of RGB-D images and $\pm 45.0$ yaw rotation perturbations (only in the real world as this degrades performance in simulation as we'll show in Section~\ref{sec:experiments}). 
%\rebuttal{The cropping operation is performed on aligned RGB and depth frames together, thus maintain pixel-level correspondence.} 
We use a batch size 16 on a Nvidia 32GB V100 GPU for 200k steps (one day) for single-task experiments, and a batch size 48 on 8 Nvidia 32GB V100 GPUs for 600K steps (5 days) for language-conditioned multi-task experiments. \rebuttal{At test time, we call upon a low-level motion planner to reach predicted keyposes. In simulation, we use native motion planner implementation provided in RLBench, which is a sampling-based BiRRT \citep{kuffner2000rrt} motion planner powered by Open Motion Planning Library (OMPL) \cite{sucan2012open} under the hood. For real-world experiments, we use the same BiRRT planner provided by the MoveIt! ROS package \citep{coleman2014reducing}.}
please, see Appendix~\ref{sec:architecture_details} for more details.

%===============================================================================

\vspace{-0.05in}
\section{Experiments}
\label{sec:experiments}
%\vspace{-0.05in}
\begin{figure}
    % \centering
    \begin{adjustbox}{center}
    \includegraphics[width=1.0\textwidth,keepaspectratio]{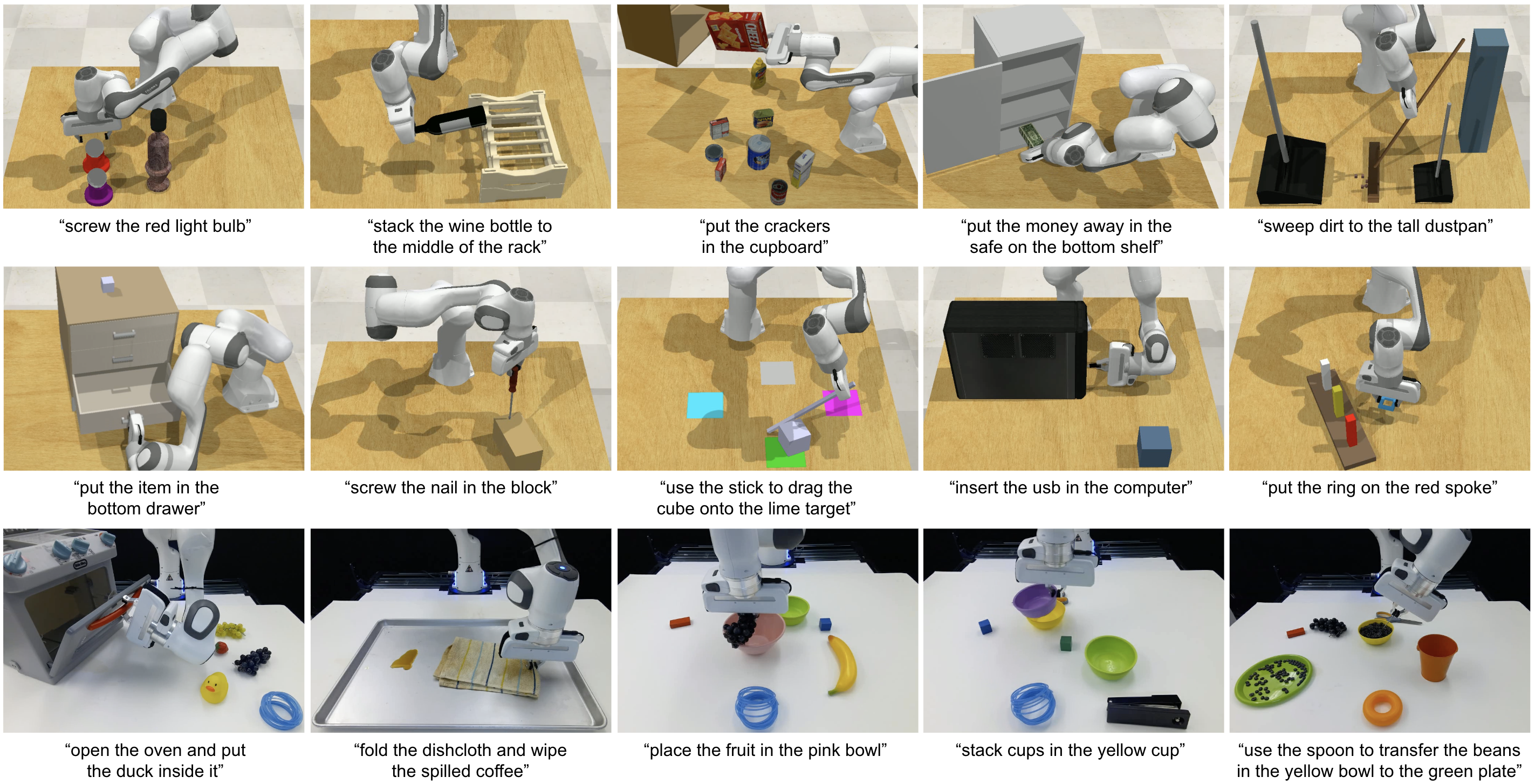}
    \end{adjustbox}
    \caption{\textbf{Tasks.} We conduct experiments on 92 simulated tasks in RLBench \cite{james2020rlbench} (only 10 shown), and 8 real-world tasks (only 5 shown). }
    \label{fig:tasks}
\end{figure}

We test \model{} in learning from demonstrations single-task and multi-task manipulation policies in simulation and the real world. 
%In the multi-task setting, task and goal conditioning are given as input through language instructions.
We conduct our simulated experiments in  RLBench~\cite{james2020rlbench}, an established simulation benchmark for learning manipulation policies, for the sake of reproducibility and benchmarking. 
%We chose RLBench~\cite{james2020rlbench} for its diversity of 6 DoF manipulation tasks and ease of generating demonstrations with templated language goals. 
Our experiments aim to answer the following questions:

\textbf{1.} How does \model{} compare against SOTA 2D multiview and 3D manipulation policies in  single-task and multi-task settings with varying number of training demonstrations?

%\textbf{2.} How does the test performance change with varying number of training demonstrations?
\textbf{2.} How does \model{} generalize across camera viewpoints compared to prior 2D multiview policies?

\textbf{3.} How do design choices such as relative 3D attention, pre-trained 2D backbones, weight-tied attention layers, and the number of coarse-to-fine sampling stages impact performance?

% \begin{enumerate}
% \item How does \model{} compare against SOTA 2D multiview and 3D manipulation policies in  single-task and multi-task settings?
% \item How does the test performance change with varying number of training demonstrations in comparison to the existing SOTA?
% \item How does \model{} generalize across camera viewpoints in comparison to existing 2D multiview policies?
% \item How do design choices such as relative 3D attention, pre-trained 2D backbones, weight-tied attention layers, and the number of coarse-to-fine sampling stages impact performance?
% \end{enumerate}

%We perform experiments to answer the following questions: (1) How effective is \model{} ~compared to SOTA 2D and 3D manipulation policies? (2) Can \model{}~be trained on real-world tasks with noisy data? (3) What components of \model{}~matter most and what are failure cases?
%\vspace{-0.05in}

\begin{figure}[t!]
    % \centering
    % \includegraphics[height=3.4cm,keepaspectratio]{final_figures/results_single_task}
    \begin{adjustbox}{center}
    \includegraphics[width=1.1\textwidth,keepaspectratio]{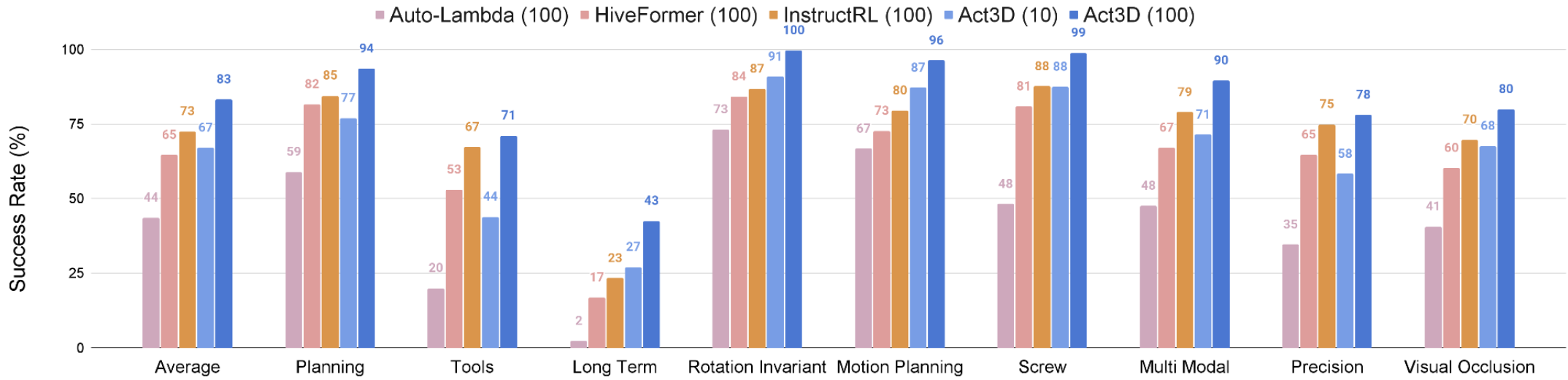}
    \end{adjustbox}
    \caption{\textbf{Single-task performance.} On 74 RLBench tasks across 9 categories, \model{}~reaches 83\% success rate, an absolute improvement of 10\% over InstructRL~\cite{liu2022instruction}, prior SOTA in this setting.}
    \label{fig:results_single_task}
\vspace{0.5cm}
     \begin{adjustbox}{center}
     \begin{subfigure}[c]{0.35\textwidth}
         \includegraphics[width=0.6\textwidth]{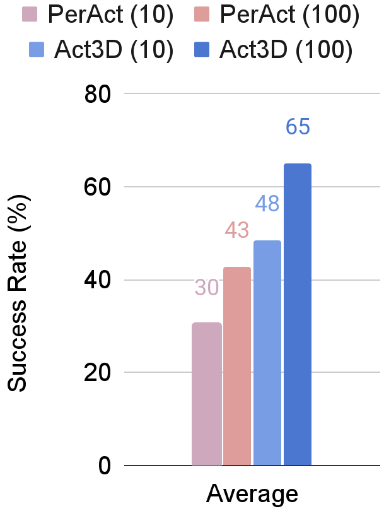}
     \end{subfigure}
     \hspace{-60pt}
     \begin{subfigure}[c]{0.8\textwidth}
         \includegraphics[width=\textwidth]{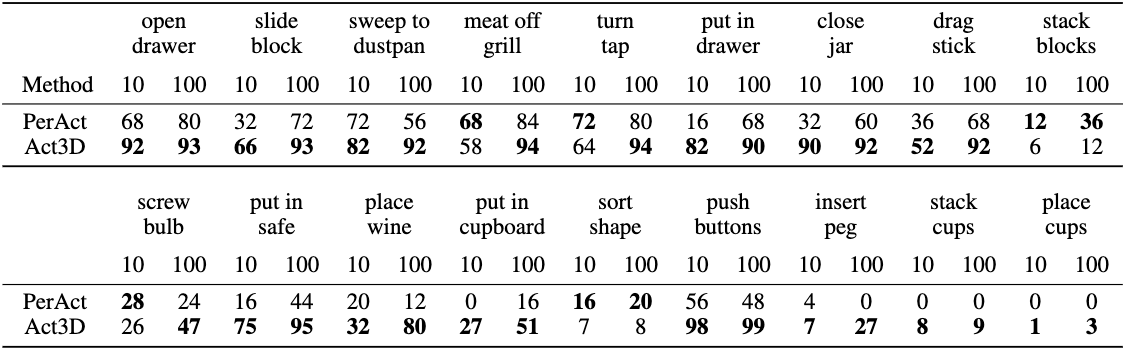}
     \end{subfigure}
    \end{adjustbox}
     \caption{\textbf{Multi-task performance.} On 18 RLBench tasks with 249 variations, \model{} reaches 65\% success rate, an absolute improvement of 22\% over PerAct~\cite{shridhar2023perceiver}, prior SOTA in this setting.}
     \label{fig:multi_task_results}
\end{figure}
\subsection{Evaluation in simulation}
%\xian{One concern here is the experiment section seems a bit too short to me, while the intro and related work are too long. Here we are merely reporting numbers without analysis. Maybe one thing ew can do is to move viewpoint change from ablations to here, and have a more detailed comparison}

\paragraph{Datasets}  We test \model{} in RLbench~in two settings: % to ensure a clear comparison with prior work: 
\textbf{1.} \textbf{Single-task} manipulation policy learning. %setting with 74 tasks proposed by HiveFormer~\cite{guhur2023instruction}. 
We consider 74 tasks grouped into 9 categories proposed by HiveFormer~\cite{guhur2023instruction}. Each task includes   variations which test generalization  
%to new scene  configurations %specified by language instructions 
%that involve 
to novel arrangements of the same training objects.
Each method is trained with 100 demonstrations and evaluated on 500 unseen episodes.  \textbf{2.}  \textbf{Multi-task} manipulation policy learning. %A multi-task multi-variation setting with 18 tasks and 249 variations proposed by PerAct~\cite{shridhar2023perceiver}. 
We consider 18 tasks with 249 variations proposed by PerAct~\cite{shridhar2023perceiver}.
Each task includes  2-60 variations, which test generalization  to new goal configurations %specified by language instructions 
that involve  novel object colors, shapes, sizes, and categories. %, over what it was seen in the training set. 
This is a more challenging setting. %, since the previous setting only tested generalization to novel arrangements of the same objects. 
Each method is trained with 100 demonstrations per task split across variations, and evaluated on 500 unseen episodes per task.
%More details on the datasets considered can be found  in the Appendix~\ref{sec:tasks}.

%\vspace{-0.02in}
\paragraph{Baselines} We compare \model{}~with the following state-of-the-art manipulation policy learning methods: 
\textbf{1.} InstructRL~\cite{liu2022instruction}, a 2D policy that directly predicts 6 DoF poses from image and language conditioning with a pre-trained vision-and-language backbone. 
\textbf{2.}  PerAct~\cite{shridhar2023perceiver},  a 3D policy that voxelizes the workspace and detects the next best voxel action through global self-attention. 
\textbf{3.} HiveFormer~\cite{guhur2023instruction} and Auto-$\lambda$~\cite{liu2022auto}, hybrid methods that detect a contact point within an image input, then regress an offset from this contact point. We report numbers from the papers when available.

%\vspace{-0.02in}
\paragraph{Evaluation metric} 
We evaluate policies by task completion success rate, the proportion of execution trajectories that lead to goal conditions specified in language instructions.

%\vspace{-0.02in}
\paragraph{Single-task and multi-task manipulation results} 
We show single-task quantitative results of our model and baselines in Figure~\ref{fig:results_single_task}. 
\model{} \textbf{reaches 83\% success rate, an absolute improvement of 10\% over InstructRL~\cite{liu2022instruction}, prior SOTA in this setting}, and consistently outperforms it across all 9 categories of tasks.
With only 10 demonstrations per task, \model{} is competitive with prior SOTA using 100 demonstrations per task. 
%\rebuttal{
Act3D outperforms 2D methods of InstructRL and Hiveformer because it reasons directly  in 3D. 
%and uses geometric operations to link 2.5D input to 3D feature space. 
For the same reason, it generalizes much better than them to novel camera placements, as we show in Table \ref{table:ablations}. %} 

We show multi-task quantitative results of our model and PerAct in Figure~\ref{fig:multi_task_results}.
\model{} reaches 65\% success rate, an absolute improvement of 22\% over PerAct, prior SOTA in this setting, consistently outperforming it across most tasks. 
\textbf{With only 10 demonstrations per task, \model{} outperforms PerAct using 100 demonstrations per task.}
Note that \model{} also uses less than a third of PerAct's training computation budget: PerAct was trained for 16 days on 8 Nvidia V100 GPUs while we train for 5 days on the same hardware. Act3D outperforms PerAct because its coarse-to-fine relative attention based 3D featurization of the 3D workspace is more effective than the perceiver’s latent bottleneck attention in generating spatially  disentangled features.

%\vspace{-0.05in}
\subsection{Ablations}
\label{sec:ablations}
%\vspace{-0.05in}
% We ablate design choices of \model{}.  
% We perform most ablations in the single-task setting on 5 tasks: pick cup, put knife on chopping board, put money in safe, slide block to target, take umbrella out of stand; %ablate whether to predict if the motion planner needs to avoid collisions to reach of the pose on all 74 tasks;
% and ablate the choice of pre-trained 2D backbone in the multi-task setting with all 18 tasks. 

We ablate the impact of our design choices in Table~\ref{table:ablations}. We perform most ablations in the single-task setting on 5 tasks: pick cup, put knife on chopping board, put money in safe, slide block to target, take umbrella out of stand. We ablate the choice of pre-trained 2D backbone in the multi-task setting with all 18 tasks.
%We show in Table~\ref{table:ablations} that the the 3D action detection inductive bias of \model{} increases robustness to viewpoint changes in comparison to hiveformer, whose  in Table~\ref{table:ablations}. 

\begin{table}[]
\caption{\textbf{Ablations.}}
\label{table:ablations}
%\begin{adjustwidth}{-.2in}{-.2in} 
%\scriptsize
\small
\centering
\begin{tabular}{llc}
\toprule
\multicolumn{1}{c}{}                        & \multicolumn{1}{c}{}                               & Average success rate in       \\
\multicolumn{1}{c}{}                        & \multicolumn{1}{l}{}                          & single-task setting (5 tasks) \\ \midrule
\multirow{6}{*}{Core design choices}        & Full Act3D   & \textbf{98.1}          \\
                                            & Only 2 stages of coarse-to-fine sampling          & 93.6         \\
                                 %           & full workspace, 16 cm ball, regress an offset      &                               \\
                                            & No weight tying across stages                      & 80.6                          \\
                                            & Absolute 3D positional embeddings                  & 55.4                          \\
                                            & Attention to only global coarse visual features    & 89.8              \\
                                            & Only 1000 ghost points at inference time           & 93.2                          \\ \midrule
\multirow{2}{*}{Viewpoint changes}          & Act3D  & \textbf{74.2}      \\
                                            & HiveFormer                                         & 20.4                          \\ \midrule
%\multirow{2}{*}{Augmentations}              & No image augmentations                             & \textbf{91.6}                          \\
%                                            & With rotation augmentations                        & 86.2                          \\ \midrule                                      
%\multirow{4}{*}{Hyperparameter sensitivity} & Double sampling ball diameters: 32 cm and 8 cm     & 96.6                          \\
%                                            & Halve sampling ball diameters: 8 cm and 2 cm       & 91.2                          \\
%                                            & 500 ghost points at training time                  & 95.8                          \\
%                                            & 2000 ghost points at training time (need 2 GPUs)   & \textbf{98.4}                 \\ \bottomrule
%                                             &                                                    &                               \\
%                                             &                                                    & Single-task setting (74 tasks)  \\ \midrule
% \multirow{2}{*}{Motion planning}            & With collision avoidance prediction                & \textbf{83.0}                 \\
%                                             & Without collision avoidance prediction             & 77.9                          \\ \bottomrule

%                                            &                                                    &                               \\
          %                                  &     
       &     & Multi-task setting (18 tasks) \\ \midrule
\multirow{3}{*}{Backbone}                   & CLIP ResNet50 backbone                             & \textbf{65.1}                 \\
                                            & ImageNet ResNet50 backbone                         & 53.4                          \\
    %                                        & No backbone (raw RGB)                              & 45.2                          \\ \bottomrule
%\multirow{3}{*}{Cameras}                    & 4 cameras: left and right shoulders, front, wrist  & 65.1                          \\
%                                            & 3 cameras: left and right shoulders, wrist         & 56.3                          \\ \bottomrule
                                            
\end{tabular}
%\end{adjustwidth}
\end{table}

%\todo{BOLD best results in table ablations}

\paragraph{Generalization across camera viewpoints:}  We vary camera viewpoints at test time for both \model{} and HiveFormer~\cite{guhur2023instruction}.  The success rate drops to 20.4\% for HiveFormer, a relative 77\% drop, while \model{} achieves 74.2\% success rate, a 24\% relative drop. This shows detecting actions in 3D makes \model{} more robust to camera viewpoint changes than multiview 2D methods that regress offsets. %To verify this hypothesis,%\todo{THEO: CAN WE MEASURE RELATIVE DROPS OF SUCCESS RATES? IT }

\paragraph{Weight-tying and coarse-to-fine sampling:} All 3 stages of coarse-to-fine sampling are necessary: a model with only 2 stages of sampling and regressing an offset from the position selected at the second stage suffers a 4.5\% performance drop. Tying weights across stages and relative 3D positional embeddings are both crucial; we observed severe overfitting without, reflected in respective 17.5\% and 42.7\% performance drops.
Fine ghost point sampling stages should attend to local fine visual features with precise positions: all stages attending to global coarse features leads to a 8.3\% performance drop. 
\model{} can effectively trade off inference computation for performance: sampling 10,000 ghost points, instead of the 1,000 the model was trained with, boosts performance by 4.9\%.

\paragraph{Pre-training 2D features:} We investigate the effect of the pre-trained 2D backbone in the multi-task setting where language instructions are most needed. A ResNet50~\cite{radford2021learning} backbone pre-trained with CLIP improves success rate by 8.7\% over a ResNet50 backbone pre-trained on ImageNet. %, and by 16.9\% over using raw RGB as the visual token features. %\model{} This shows that enabling easily leveraging pre-trained 2D backbones while detecting actions in 3D is an important benefit of the \model{} framework.

We found Random crops of RGB-D images to boost performance but yaw rotation perturbations did not help. The model  is robust to variations in hyperparameters such as the diameter of ghost point sampling balls  or the number of points sampled during training. For additional ablations regarding augmentations and sensitivity to hyperparameters, please see the Appendix section \ref{sec:ablateappendix}.

\subsection{Evaluation in real-world}

\begin{wraptable}{r}{0.35\textwidth}
  \setlength\tabcolsep{2.3pt}
  \centering
  % \small
\begin{tabular}{lcc} 
\toprule
Task          & \# Train~  & Success  \\ 
\midrule
reach target &10        & 10/10        \\
duck in oven &15           & 6/10        \\
wipe coffee   &15           & 7/10        \\
fruits in bowl & 10          & 8/10        \\
stack cups & 15            & 6/10        \\
transfer beans  & 15           & 5/10        \\
press handsan  & 10            & 10/10        \\
uncrew cap   & 10              & 8/10        \\
\bottomrule
\end{tabular}
    \caption{Real-world tasks.}
  % \vspace{-3em}
  \label{tab:real}
\end{wraptable}In our real-world setup, we conduct experiments with a Franka Emika Panda robot and a single Azure Kinect RGB-D sensor. %; more details are in Appendix~\ref{sec:realworld}.
We consider 8 tasks (Figure \ref{fig:tasks}) that involve interactions with multiple types of objects, spanning liquid, articulated objects, and deformable objects. 
For each task, we collected 10 to 15 kinesthetic demonstrations
%, teleoperated using a 6-DoF SpaceMouse Compact\footnote{\url{https://3dconnexion.com/us/product/spacemouse-compact/}} controller, 
and trained a languaged-conditioned multi-task model with all of them. 
We report the success rate on 10 episodes per task in Table \ref{tab:real}. 
\model{} can capture semantic knowledge in demonstration well and performs reasonably well on all tasks, even with a single camera input. 
One major failure case comes from noisy depth sensing: when the depth image is not accurate, the selected point results in imprecise action prediction. 
Leveraging multi-view input for error correction could improve this, and we leave this for future work. For   videos of the robot executing the tasks, please see our project website.

%\noindent \textbf{Motion planning:} We verify that it is necessary for \model{} to predict whether the motion planner should avoid collisions to reach the predicted keypose: performance drops by 5.1\% without. Note that this is the case for all baselines that rely on a motion planner to reach keyposes, such as state-of-the-art methods of HiveFormer \cite{guhur2023instruction} and PerAct \cite{shridhar2023perceiver}. 

%For video results of \model{} in simulation and in the real world, please see the supplementary file. 

%\subsection{Limitations - future work}
%\vspace{-0.05in}
\subsection{Limitations and future work}
Our framework currently has the following limitations: 
%\textbf{1.} 
%\model{} sometimes fails in very high-precision tasks, like screwing and insertions, requiring temporally fine-grain closed-loop control. 
\textbf{1.} 
\model{} is limited by the motion planner used to connect predicted  keyposes with straight trajectory segments. It does not handle manipulation of articulated object  well, such as opening/closing doors, fridges, and ovens, where robot trajectories cannot be well approximated by few line segments.%which require a more precise trajectory than the one supplied by a motion planner that connects predicted  keyposes with straight trajectory segments.
%Learning-based trajectory prediction~\cite{fishman2023motion, chi2023diffusion} would help. %, bypassing the need to predict whether the planner should avoid collisions to reach the next keypose, as currently done by many keypose prediction frameworks \cite{shridhar2023perceiver}.  
\textbf{2.} 
\model{}  does not utilize any  decomposition of tasks into subtasks.
%, rather predicts a fla predicts keyposes one by one, without any 
%Currently, for long horizon tasks our policy would need to predict all keyposes one by one. 
A hierarchical framework that would predict language subgoals for subtasks~\cite{ahn2022can, huang2023grounded, lin2023text2motion} and feed those to our language-conditioned policy would allow better re-usability of skills across tasks. %with LLM task planning %as well as object-centric tokenization~\cite{jiang2022vima, stone2023open}, would help. 
%A hierarchical framework, where the language descriptions would be used to guide intermediate manipulation actions, would allow the re-usability of skills across tasks.
%All keypose prediction methods share the listed limitations. 
Addressing these limitations is a direct avenue for future work. % as they are orthogonal to our framework.

%==============================================================================

\vspace{-0.05in}
\section{Conclusion}
\label{sec:conclusion}
%\vspace{-0.05in}
We presented \model{}, a language-conditioned policy transformer that predicts continuous resolution 3D action maps for multi-task robot manipulation. \model{} represents the scene using a continuous resolution 3D feature map, obtained by coarse-to-fine 3D point sampling and attention-based featurization. 
%architecture that learns manipulation policies from demonstrations. 
%From one or more posed RGB-D images and language instructions, it predicts 6-DoF robot end-effector keyposes by iteratively selecting and featurizing 3D point grids in the robot's  workspace. 
%We showed \model{} %The proposed learning-based active featurization of the empty 3D space 
%is much more computationally efficient than  approaches that  exhaustively featurize all 3D voxels. 
%bypasses exhaustive processing of all 3D voxels in the workspace, and uses flexible transformer architecture with relative position attention for generalization across scene rearrangements. %to handle the arbitrariness of world coordinate system voxelization 
%with known camera extrinsics and intrinsics  on language instructions 
\model{} sets a new state-of-the-art in RLBench, an established robot manipulation benchmark, and solves diverse manipulation tasks in the real world from a single RGB-D camera view and a handful of demonstrations. 
Our ablations quantified the contribution of relative 3D attentions, 2D feature pre-training, and weight tying during coarse-to-fine iterations. %Extending \model{} to handle more precise tasks, and tasks involving trajectory-level action prediction remain as our future work.
%empiricallyt  show our lations tha3D inductive bias, 2D feature pre-training, relativev  helps the policy  generalize across viewpoints better than existing 2D or 3D  baselines,  helps and 
%We hope our model to serve as a general baseline for learning manipulation from demonstrzation thanks to its simplicity, generality and sample efficiency. 
%We will make our code and models available to the community to ensure reproducibility.
%and and ablate design choices regarding weight tying and relative attention, as important for good performance. 
%\todo{a comment about 2d features..}
%We hope our model can serve as a simple basel 

%a language-conditioned manipulation policy architecture that uses recu

%===============================================================================

\section{Acknowledgements} This work is supported by Sony AI, NSF award No 1849287, DARPA Machine Common Sense, an Amazon faculty award, and an NSF CAREER award.

%===============================================================================

\bibliography{references}  % .bib

\newpage
\section{Appendix}

\subsection{Real-world Setup}
\label{sec:realworld}

\begin{wrapfigure}{r}{0.5\textwidth}
    \centering
    \includegraphics[width=0.5\textwidth,keepaspectratio]{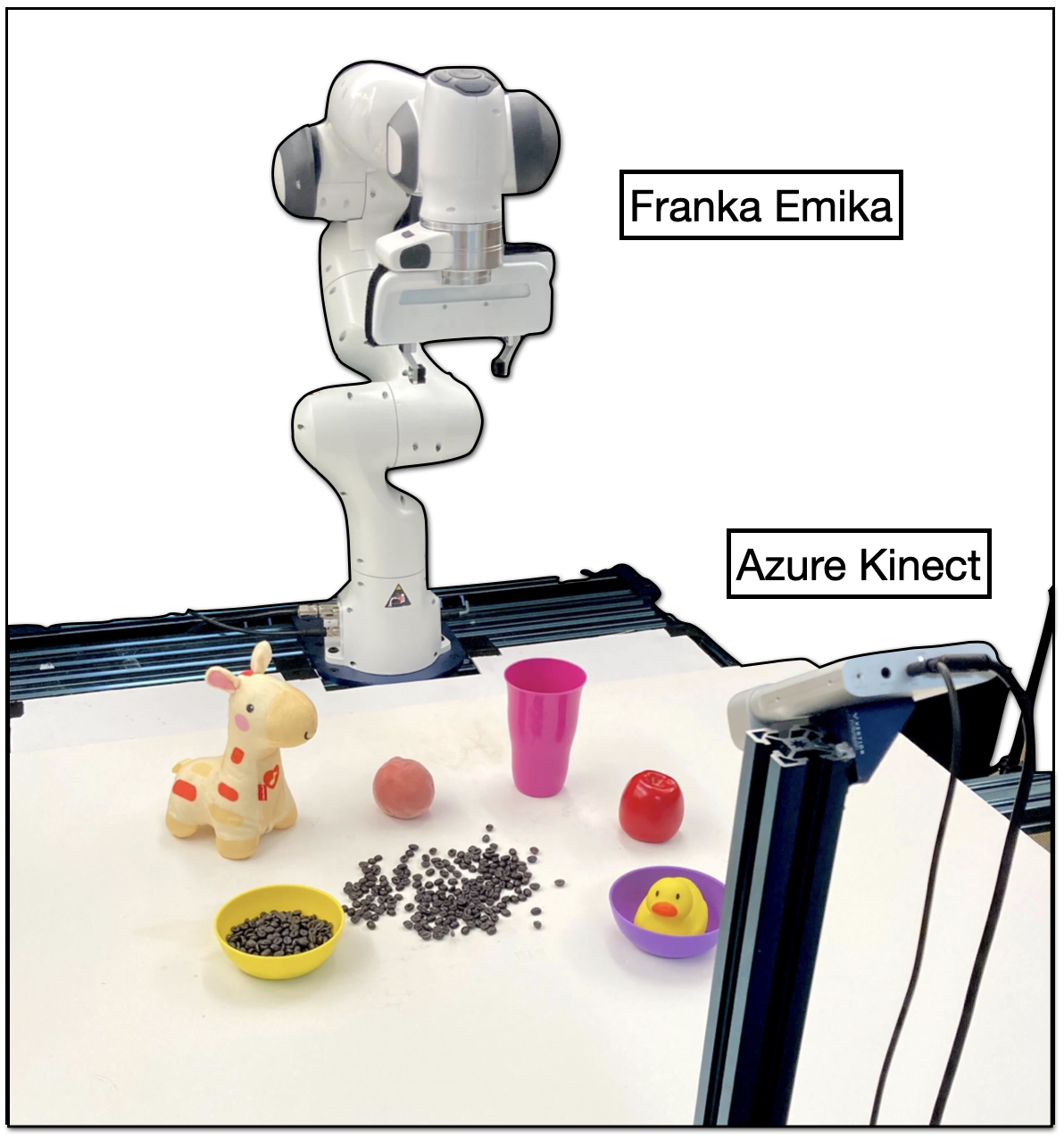}
    \caption{\textbf{Real-world setup.}}
    \label{fig:robot}
    \vspace{1cm}
    \includegraphics[width=0.5\textwidth,keepaspectratio]{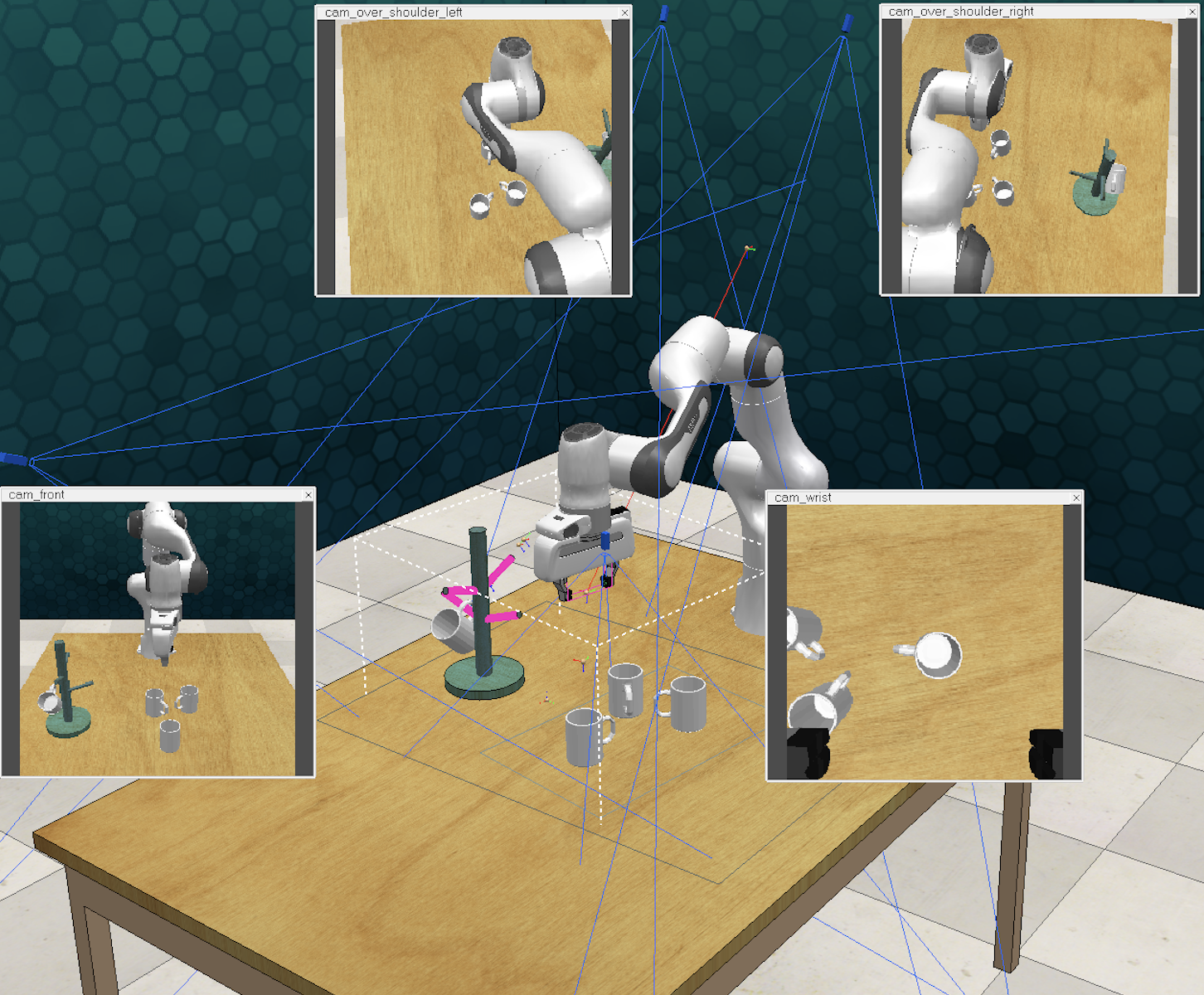}
    \caption{\textbf{RLbench simulation setup.}}
    \label{fig:sim}
    \vspace{-15em}
\end{wrapfigure}

Our real-robot setup contains a Franka Panda robotic arm equipped with a parallel jaw gripper, as shown in Figure \ref{fig:robot}. 
We get RGB-D input from a single Azure Kinect sensor at a front view at 30Hz. 
The image input is of resolution $1280\times 720$, we crop and downsample it to $256\times256$. 
We calibrate the extrinsics of the camera with respect to the robot base using the \texttt{easy\_handeye}\footnote{\url{https://github.com/IFL-CAMP/easy_handeye}} ROS package. 
We extract keyposes from demonstrations in the same was as in simulation.
Our real-world multi-task policy is trained on 4 V100 GPUs for 3 days, and we run inference on a desktop with a single RTX4090 GPU. 
For robot control, we use the open-source \texttt{frankapy}\footnote{\url{https://github.com/iamlab-cmu/frankapy}} package to send real-time position-control commands to the robot.

\subsection{RLBench Simulation Setup}
\label{sec:sim}

To ensure fair comparison with prior work, we use $n_{\textrm{\small cam}} \in \{3, 4\}$ cameras for simulated experiments depending on the evaluation setting. In our single-task evaluation setting first proposed by HiveFormer~\cite{guhur2023instruction}, we use the same 3 cameras they do $\{O_{\textrm{\small left}}, O_{\textrm{\small right}}, O_{\textrm{\small wrist}}\}$. In our multi-task evaluation setting first proposed by PerAct~\cite{shridhar2023perceiver}, we use the same 4 cameras they do $\{O_{\textrm{\small front}}, O_{\textrm{\small left}}, O_{\textrm{\small right}}, O_{\textrm{\small wrist}}\}$.

\newpage
\subsection{RLBench Tasks}
\label{sec:tasks}

\begin{figure}[h!]
    \centering
    \begin{adjustbox}{center}
\includegraphics[width=1.1\textwidth,keepaspectratio]{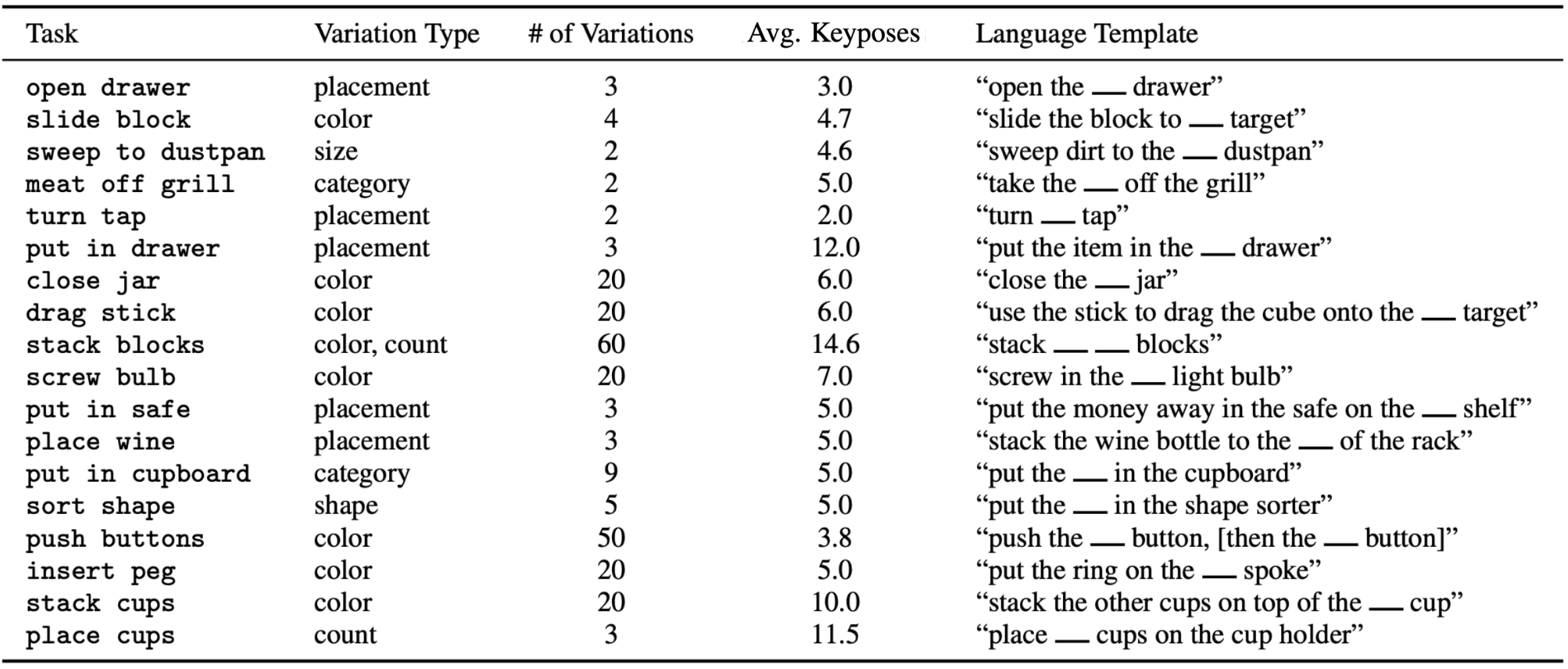}
    \end{adjustbox}    
    \caption{\textbf{PerAct~\cite{shridhar2023perceiver} tasks.} 
    We adopt the multi-task multi-variation setting from PerAct~\cite{shridhar2023perceiver} with 18 tasks and 249 unique variations across object placement, color, size, category, count, and shape.
    }
    \label{fig:peract_tasks}
\end{figure}

We adapt the single-task setting of HiveFormer~\cite{guhur2023instruction} with 74 tasks grouped into 9 categories according to their key challenges. The 9 task groups are defined as follows:
\begin{itemize}
    \item The \textbf{Planning} group contains tasks with multiple sub-goals (e.g. picking a basket ball and then
throwing the ball). The included tasks are: basketball in hoop, put rubbish in bin, meat off grill,
meat on grill, change channel, tv on, tower3, push buttons, stack wine.
    \item The \textbf{Tools} group is a special case of planning where a robot must grasp an object to interact
with the target object. The included tasks are: slide block to target, reach and drag, take frame
off hanger, water plants, hang frame on hanger, scoop with spatula, place hanger on rack, move
hanger, sweep to dustpan, take plate off colored dish rack, screw nail.
    \item The \textbf{Long term} group requires more than 10 macro-steps to be completed. The included tasks
are: wipe desk, stack blocks, take shoes out of box, slide cabinet open and place cups.
    \item The \textbf{Rotation-invariant} group can be solved without changes in the gripper rotation. The
included tasks are: reach target, push button, lamp on, lamp off, push buttons, pick and lift,
take lid off saucepan.
    \item The \textbf{Motion planner} group requires precise grasping. As observed in [81] such tasks often fail
due to the motion planner. The included tasks are: toilet seat down, close laptop lid, open box,
open drawer, close drawer, close box, phone on base, toilet seat up, put books on bookshelf.
    \item The \textbf{Multimodal} group can have multiple possible trajectories to solve a task due to a large
affordance area of the target object (e.g. the edge of a cup). The included tasks are: pick up
cup, turn tap, lift numbered block, beat the buzz, stack cups.
    \item The \textbf{Precision} group involves precise object manipulation. The included tasks are: take usb
out of computer, play jenga, insert onto square peg, take umbrella out of umbrella stand, insert usb in computer, straighten rope, pick and lift small, put knife on chopping board, place shape
in shape sorter, take toilet roll off stand, put umbrella in umbrella stand, setup checkers.
    \item The \textbf{Screw} group requires screwing an object. The included tasks are: turn oven on, change
clock, open window, open wine bottle.
    \item The \textbf{Visual Occlusion} group involves tasks with large objects and thus there are occlusions
from certain views. The included tasks are: close microwave, close fridge, close grill, open
grill, unplug charger, press switch, take money out safe, open microwave, put money in safe,
open door, close door, open fridge, open oven, plug charger in power supply
\end{itemize}

\subsection{Further Architecture Details}
\label{sec:architecture_details}

\paragraph{Relative 3D cross-attentions} 
We featurize each of the 3D ghost points and a parametric query (used to select via inner-product one of the ghost points as the next best end-effector position in the decoder) independently through cross-attentions to the multi-scale 3D scene feature cloud, language tokens, and proprioception.
Featurizing ghost points independently, without self-attentions to one another, enables sampling more ghost points at inference time to improve performance, as we show in Section~\ref{sec:experiments}. 
Our cross-attentions use relative 3D position information and are implemented efficiently with rotary positional embeddings~\cite{su2021roformer}.

Given a point $\mathbf{p} = (x, y, z) \in \mathbb{R}^3$ and its feature $\mathbf{x} \in \mathbb{R}^d$, the rotary position encoding function $\mathbf{PE}$ is defined as:

\begin{align}
\mathbf{PE}(\mathbf{p}, \mathbf{x}) 
    = \mathbf{M}(\mathbf{p}) \mathbf{x} 
    = \begin{bmatrix}
        \mathbf{M}_1 &        &           \\
           & \ddots &           \\
           &        & \mathbf{M}_{d / 6} \\
      \end{bmatrix} \mathbf{x}
\end{align}

\begin{align}
\mathbf{M}_k
    = \begin{bmatrix}
        \cos x \theta_k & -\sin x \theta_k & 0 & 0 & 0 & 0 \\
        \sin x \theta_k & \cos x \theta_k  & 0 & 0 & 0 & 0 \\
        0                & 0                 & \cos y \theta_k & -\sin y \theta_k & 0 & 0 \\
        0                & 0                 & \sin y \theta_k & \cos y \theta_k  & 0 & 0 \\
        0                & 0                 & 0                & 0                 & \cos z \theta_k & -\sin z \theta_k \\
        0                & 0                 & 0                & 0                 & \sin z \theta_k & \cos z \theta_k \\
      \end{bmatrix} 
\end{align}

where $\theta_k = \frac{1}{10000^{6(k - 1) / d}}$, \rebuttal{$k \in \{1, .., d/6\}$}. The dot product of two positionally encoded features is
\begin{align}
\mathbf{PE}(\mathbf{p}_i, \mathbf{x}_i)^T \mathbf{PE}(\mathbf{p}_j, \mathbf{x}_j)
     = \mathbf{x}_i^T \mathbf{M}(\mathbf{p}_i)^T \mathbf{M}(\mathbf{p}_j) \mathbf{x}_j
     = \mathbf{x}_i^T \mathbf{M}(\mathbf{p}_j - \mathbf{p}_i) \mathbf{x}_j
\end{align}
which depends only on the relative positions of points $\mathbf{p}_i$ and $\mathbf{p}_j$.

%\paragraph{Implementation details}
We extract two feature maps per $256$x$256$ input image view: $32$x$32$ coarse visual tokens and $128$x$128$ fine visual tokens.
We use three ghost point sampling stages: first \rebuttal{uniformly} across the entire workspace (roughly $1$ meter cube), then \rebuttal{uniformly} in a $16$ centimeter diameter ball, and finally in a $4$ centimeter diameter ball. 
The coarsest ghost points attend to a global coarse scene feature cloud ($32$x$32$x$n_{\textrm{\small cam}}$ coarse visual tokens) whereas finer ghost points attend to a local fine scene feature cloud (the closest $32$x$32$x$n_{\textrm{\small cam}}$ out of the total $128$x$128$x$n_{\textrm{\small cam}}$ fine visual tokens). 
During training, we sample $1000$ ghost points in total split equally across the three stages. 
At inference time, we can trade-off extra prediction precision and task performance for additional compute by sampling more ghost points than the model ever saw at training time ($10,000$ in our experiments). 
We'll show in ablations in Section~\ref{sec:experiments} that our framework is robust to these hyper-parameters but tying weights across sampling stages and relative 3D cross-attention are both crucial for generalization.
We use $2$ layers of cross-attention and an embedding size $60$ for single-task experiments and $120$ for multi-task experiments.
Training samples are augmented with random crops of RGB-D images and $\pm 45.0$ yaw rotation perturbations (only in the real world as this degrades performance in simulation as we'll show in Section~\ref{sec:experiments}). 
The cropping operation is performed on aligned RGB and depth frames together, thus maintain pixel-level correspondence.
We use a batch size 16 on a Nvidia 32GB V100 GPU for 200k steps (one day) for single-task experiments, and a batch size 48 on 8 Nvidia 32GB V100 GPUs for 600K steps (5 days) for language-conditioned multi-task experiments. \rebuttal{At test time, we call a low-level motion planner to reach predicted keyposes. In simulation, we use native motion planner implementation provided in RLBench, which is a sampling-based BiRRT \citep{kuffner2000rrt} motion planner powered by Open Motion Planning Library (OMPL) \cite{sucan2012open} under the hood. For real-world experiments, we use the same BiRRT planner provided by the MoveIt! ROS package \citep{coleman2014reducing}.}

\begin{figure}[h!]
    \centering
    \begin{adjustbox}{center}
\includegraphics[width=1.1\textwidth,keepaspectratio]{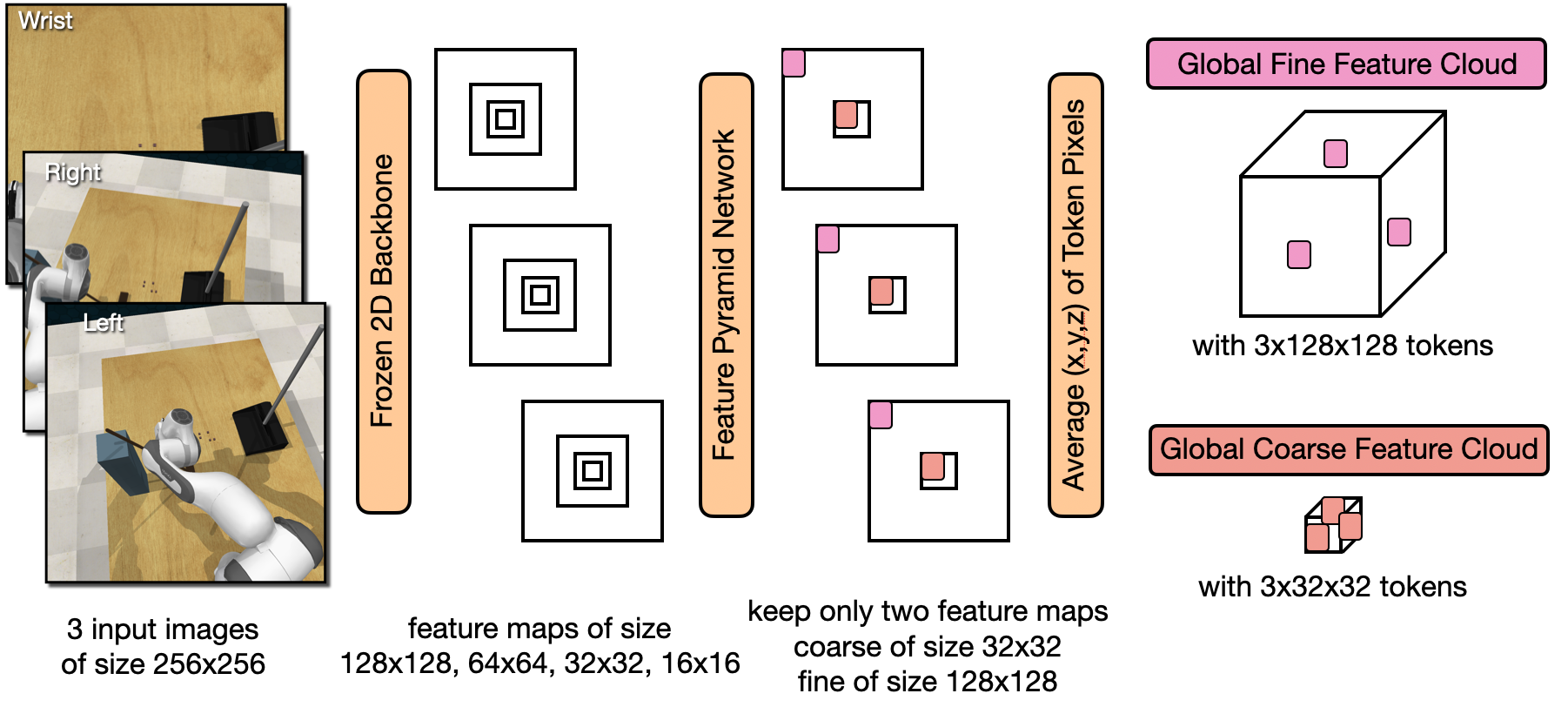}
    \end{adjustbox}    
    \caption{\textbf{Scene Feature Cloud Generation}. We encode each image independently with a pre-trained and frozen vision backbone to get multi-scale feature maps, pass these feature maps through a feature pyramid network and retain only two: a coarse feature map (at a granularity that lets ghost points attend to all tokens within GPU memory) and a fine feature map (as spatially precise as afforded by input images and the backbone). We lift visual tokens from these two feature maps for each image to 3D scene feature clouds by averaging the positions of pixels in each 2D visual token.
    }
    \vspace{-5pt}
    \label{fig:architecture1}
\end{figure}

\begin{figure}[h!]
    \centering
    \begin{adjustbox}{center}
\includegraphics[width=1.1\textwidth,keepaspectratio]{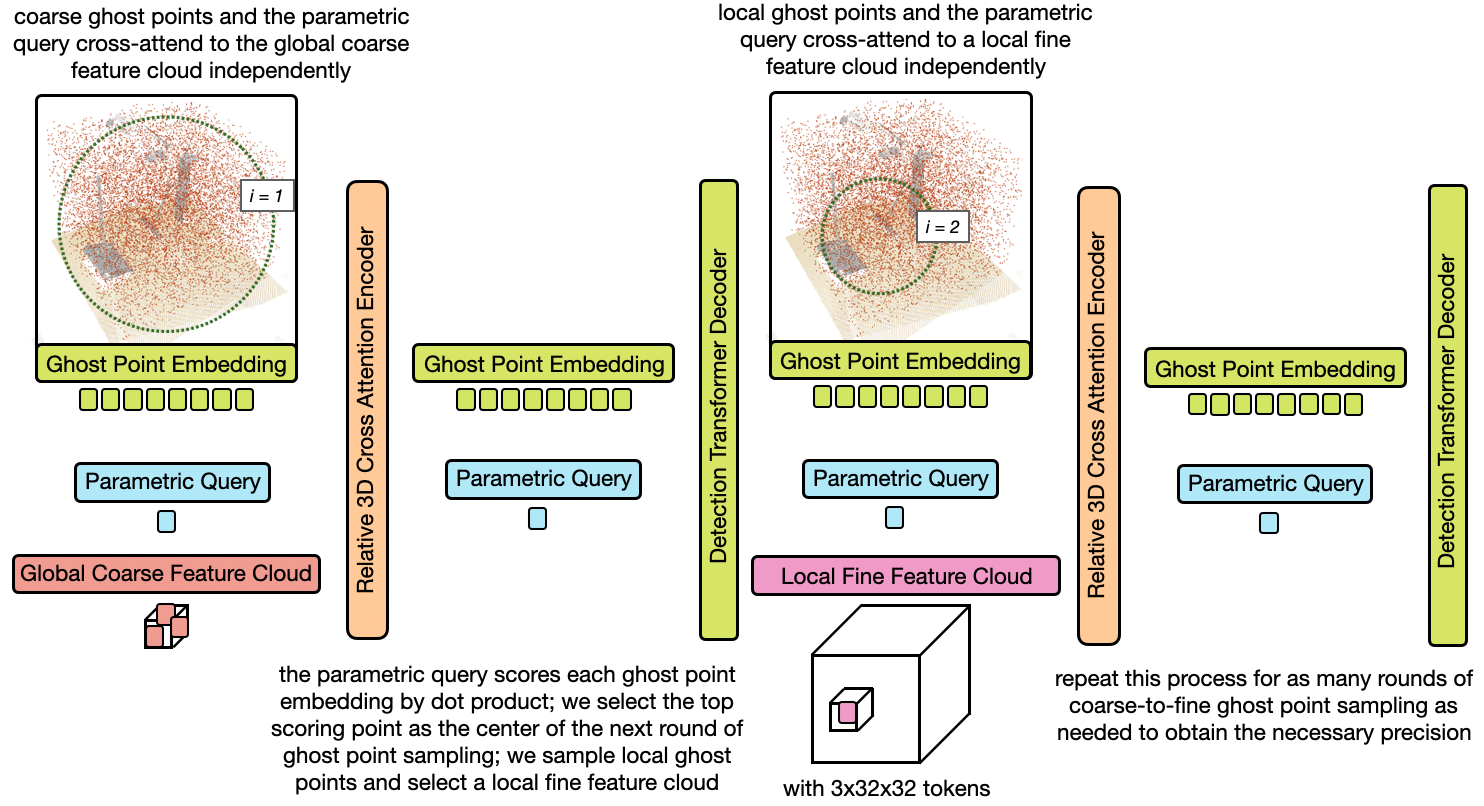}
    \end{adjustbox}    
    \caption{\textbf{Iterative Ghost Point Sampling, Featurization, and Selection}. 
    }
    \vspace{-5pt}
    \label{fig:architecture2}
\end{figure}

\begin{figure}[h!]
    \centering
    \begin{adjustbox}{center}
\includegraphics[width=1.1\textwidth,keepaspectratio]{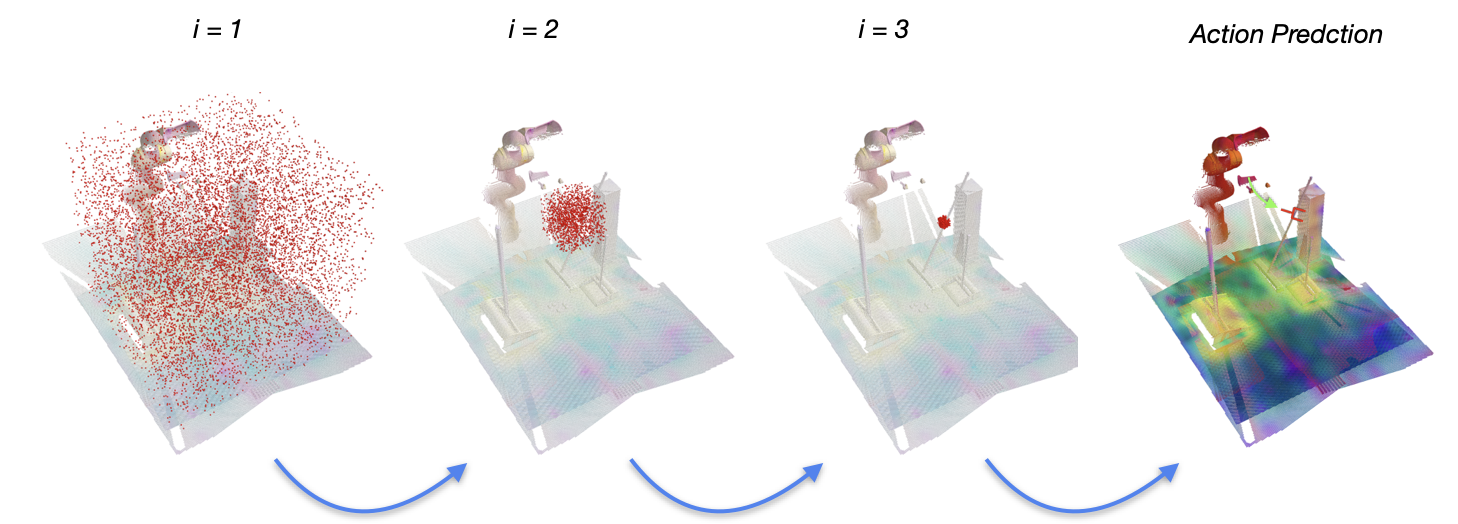}
    \end{adjustbox}    
    \caption{\textbf{Iterative Ghost Point Sampling, Featurization, and Selection}. 
    }
    \vspace{-5pt}
    \label{fig:architecture3}
\end{figure}

\rebuttal{
\subsection{High Precision Experiments}
\label{sec:high_precision_experiments}

In this section, we further investigate the ability of \model{} to improve over existing 3D methods that voxelize the workspace for high-precision tasks. 
We compare two variants of \model{} against PerAct~\cite{shridhar2023perceiver} on three high-precision tasks in success rate. 
The first \model{} variant is the standard architecture used in the remainder of our experiments operating on $256$x$256$ input image views; the second operates on higher resolution $512$x$512$ input image views, from which it extracts four times as many visual tokens with more precise 3D positions. 
This further tests the ability of \model{} to provide high precision by processing higher-resolution RGB-D views at the cost of extra compute.

\begin{table}[h]
\centering
\begin{tabular}{rccc}
\toprule
Method          & insert peg & sort shape & screw nail \\ \midrule
PerAct          & 16         & 31         & 12         \\
Act3D (256x256) & 29         & 34         & 31         \\
Act3D (512x512) & \textbf{47}         & \textbf{43}         & \textbf{55}         \\ \bottomrule
\end{tabular}
\end{table}

\model{} improves over PerAct on high precision tasks and can further benefit from higher resolution RGB-D images, at the cost of extra compute.
}

\subsection{Further ablations}
\label{sec:ablateappendix}
\begin{table}[]
\caption{\textbf{Ablations.}}
\label{table:ablations}
%\begin{adjustwidth}{-.2in}{-.2in} 
%\scriptsize
\small
\centering
\begin{tabular}{llc}
\toprule
\multicolumn{1}{c}{}                        & \multicolumn{1}{c}{}                               & Average success rate in       \\
\multicolumn{1}{c}{}                        & \multicolumn{1}{c}{Model}                          & single-task setting (5 tasks) \\ \midrule
\multirow{6}{*}{Core design choices}        & Best Act3D model (evaluated in Fig.~\ref{fig:results_single_task}) & \textbf{98.1}          \\
                                            & Only 2 stages of coarse-to-fine sampling:          & \multirow{2}{*}{93.6}         \\
                                            & full workspace, 16 cm ball, regress an offset      &                               \\
                                            & No weight tying across stages                      & 80.6                          \\
                                            & Absolute 3D positional embeddings                  & 55.4                          \\
                                            & Attention to only global coarse visual features    & 89.8              \\
                                            & Only 1000 ghost points at inference time           & 93.2                          \\ \midrule
\multirow{2}{*}{Viewpoint changes}          & Best Act3D model (evaluated in Fig.~\ref{fig:results_single_task}) & \textbf{74.2}      \\
                                            & HiveFormer                                         & 20.4                          \\ \midrule
\multirow{2}{*}{Augmentations}              & No image augmentations                             & \textbf{91.6}                          \\
                                            & With rotation augmentations                        & 86.2                          \\ \midrule                                      
\multirow{4}{*}{Hyperparameter sensitivity} & Double sampling ball diameters: 32 cm and 8 cm     & 96.6                          \\
                                            & Halve sampling ball diameters: 8 cm and 2 cm       & 91.2                          \\
                                            & 500 ghost points at training time                  & 95.8                          \\
                                            & 2000 ghost points at training time (need 2 GPUs)   & \textbf{98.4}                 \\ \bottomrule

%                                             &                                                    &                               \\
%                                             &                                                    & Single-task setting (74 tasks)  \\ \midrule
% \multirow{2}{*}{Motion planning}            & With collision avoidance prediction                & \textbf{83.0}                 \\
%                                             & Without collision avoidance prediction             & 77.9                          \\ \bottomrule

                                            &                                                    &                               \\
                                            &                                                    & Multi-task setting (18 tasks) \\ \midrule
\multirow{3}{*}{Backbone}                   & CLIP ResNet50 backbone                             & \textbf{65.1}                 \\
                                            & ImageNet ResNet50 backbone                         & 53.4                          \\
                                            & No backbone (raw RGB)                              & 45.2                          \\ \bottomrule
%\multirow{3}{*}{Cameras}                    & 4 cameras: left and right shoulders, front, wrist  & 65.1                          \\
%                                            & 3 cameras: left and right shoulders, wrist         & 56.3                          \\ \bottomrule
                                            
\end{tabular}
%\end{adjustwidth}
\end{table}

\noindent \textbf{Augmentations:} Random crops of RGB-D images boost success rate by 6.5\%, but yaw rotation perturbations drop it by 11.9\%. This is in line with PerAct~\cite{shridhar2023perceiver} results in RLBench.

\noindent \textbf{Hyperparameter sensitivity:} \model{}  is robust to variations in hyperparameters.  Doubling the diameter of ghost point sampling balls from (16 cm, 4 cm) to (32 cm, 8 cm) drops success rate by 1.5\% and halving it to (8 cm, 2 cm) by 6.9\%. Halving the total number of ghost points sampled from 1,000 to 500 drops success rate by 2.3\% whereas doubling it to 2,000 increases success rate by 0.3\%. We use 1,000 ghost points in our experiments to allow training with a single GPU per task.

\end{document}